\documentclass[10pt,twocolumn,letterpaper]{article}

\usepackage[pagenumbers]{cvpr} 

\usepackage{graphicx}
\usepackage{amsmath}
\usepackage{amssymb}
\usepackage{booktabs}

\usepackage{amsmath}
\usepackage{bm}
\usepackage{amssymb}
\usepackage{xspace}

\usepackage{booktabs}
\usepackage{multirow}
\usepackage{adjustbox}

\usepackage{pifont}
\usepackage{xcolor}

\usepackage{marvosym}
\newcommand{\new}[1]{#1}
\newcommand{\finalrev}[1]{{#1}}

\DeclareMathOperator*{\argmin}{arg\,min}

\definecolor{darkred}{RGB}{204, 0, 0}
\definecolor{darkgreen}{RGB}{0, 160, 0}

\newcommand{\ccmark}{{\color{darkgreen}\ding{52}}}%
\newcommand{\cxmark}{{\color{darkred}\ding{56}}}%

\DeclareUnicodeCharacter{20AC}{\EUR{}}

\usepackage[pagebackref,breaklinks,colorlinks]{hyperref}

\usepackage[capitalize]{cleveref}
\crefname{section}{Sec.}{Secs.}
\Crefname{section}{Section}{Sections}
\Crefname{table}{Table}{Tables}
\crefname{table}{Tab.}{Tabs.}

\def\cvprPaperID{*****} 
\def\confName{CVPR}
\def\confYear{2022}

\begin{document}

\title{Scene-Aware 3D Multi-Human Motion Capture from a Single Camera}

\author{D.\,C. Luvizon$^{1}$, M. Habermann$^1$, V. Golyanik$^1$, A. Kortylewski$^{1,2}$, C. Theobalt$^1$\\
    \\
    {
        \parbox{\textwidth}{%
            \centering $^1$MPI Informatics, Saarland Informatics Campus, Germany\\
            $^2$University of Freiburg, Germany
        }
    }
}


\twocolumn[{%
\renewcommand\twocolumn[1][]{#1}%
\maketitle
\begin{center}
    \centering
    \captionsetup{type=figure}
    \includegraphics[width=\linewidth]{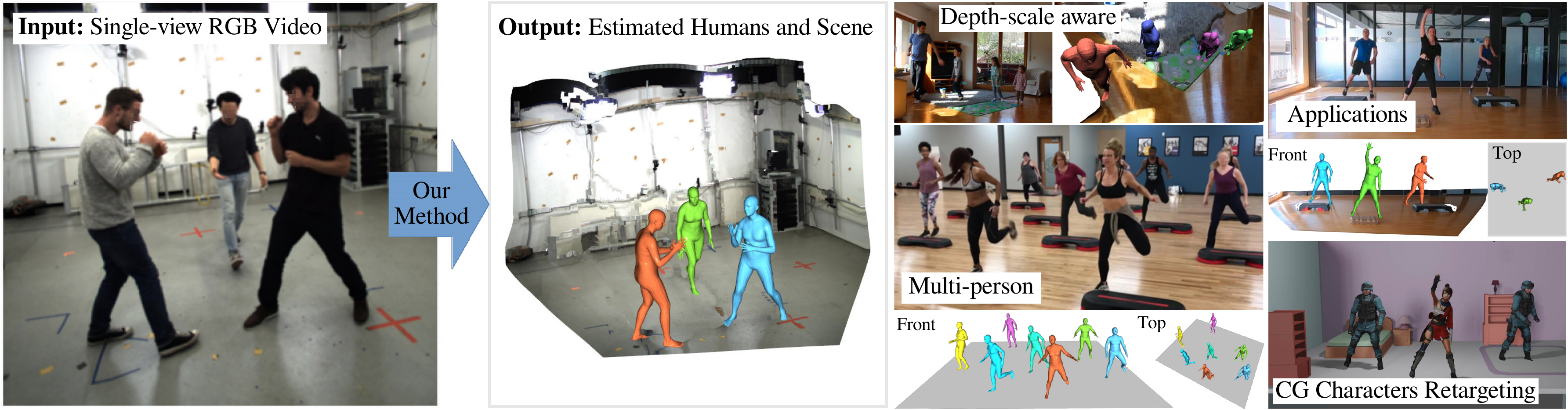}
    \captionof{figure}{
  Our approach estimates absolute 3D positions of multiple humans in a scene, body shape and articulation in a globally and temporally coherent manner from a single monocular RGB video. It 
  achieves higher 3D reconstruction accuracy than competing methods, 
  allows motion re-targeting in the 3D space,
  and works exceptionally well even for in-the-wild videos.
    }
    \label{fig:teaser}
\end{center}%
}]


%
%
\begin{abstract}
In this work, we consider the problem of estimating the 3D position of multiple humans in a scene as well as their body shape and articulation from a single RGB video recorded with a static camera.
In contrast to expensive marker-based or multi-view systems, our lightweight setup is ideal for private users as it enables an affordable 3D motion capture that is easy to install and does not require expert knowledge.
\new{%
To deal with this challenging setting, we leverage recent advances in computer vision using large-scale pre-trained models for a variety of modalities, including 2D body joints, joint angles, normalized disparity maps, and human segmentation masks.
}
\new{%
Thus, we introduce the first non-linear optimization-based approach that jointly solves for the absolute 3D position of each human, their articulated pose, their individual shapes as well as the scale of the scene.
}
In particular, we estimate the scene depth and person \finalrev{unique} scale from normalized disparity predictions using the 2D body joints and joint angles.
Given the per-frame scene depth, we reconstruct a point-cloud of the static scene in 3D space.
Finally, given the per-frame \finalrev{3D} estimates of the humans and scene point-cloud, we perform a space-time coherent optimization over the video to ensure temporal, spatial and physical plausibility.
We evaluate our method on established multi-person 3D human pose benchmarks where we consistently outperform previous methods and we qualitatively demonstrate that our method is robust to in-the-wild conditions including challenging scenes with people of different sizes.
Code: \url{https://github.com/dluvizon/scene-aware-3d-multi-human}
\end{abstract}

%
%
\section{Introduction} \label{sec:intro}
%
%
Estimating the absolute 3D position, body shape, and articulation of multiple people in a scene is a fundamental research problem that has many applications in game development, VR/AR, and HCI.
Years of research went into developing sophisticated and expensive setups such as multi-view systems, motion capture suits, and manually or semi-automatically denoising of the tracked motions to then, for example, animate CG characters with these captured motions.
However, one ideally would like to obtain such an absolute scene understanding from a capture setup that is easy to install, affordable, and that does not require expert knowledge, i.e. a \textit{single RGB camera}.
Such a lightweight setup would enable 3D motion capture for private users, e.g. avatar control via the smartphone, but it can also be applied for post production in the movie industry where, for example, one person should be replaced by another in a 3D consistent manner.
At the same time, it has to be stated that performing motion capture given such limited data is exceptionally more difficult compared to multi-view systems.
The major challenges for such a monocular setting, where only a single static video of the entire scene with moving persons is given, are the inherent depth ambiguity and occlusions, among many others. 
%
%
\par
Therefore, recent monocular approaches focus on a single human~\cite{Mehta_2017_3DV, Pavlakos_2017_CVPR} or even assume an actor template is given~\cite{xu18, habermann19, habermann20}.
Recently, some works started to research the multi-person setting, but they either only learn a relative depth ordering of people in the scene~\cite{jiang2020coherent} that is not 3D consistent over time or they directly predict absolute depth, which is prone to overfit to the settings shown in the training data~\cite{Moon_2019_ICCV_3DMPPE}.
Most of those works leverage recent advances in Computer Vision and take as input several types of regressed data modalities obtained from models trained on large-scale data.
This involves 1) 2D body joints~\cite{openposeTPAMI19, fang2017rmpe}, 2) joint angles~\cite{ROMP}, 3) normalized disparity maps~\cite{ranftl2021vision, li2019learning}, and 4) human segmentation masks~\cite{cheng2021masked}.
Interestingly, none of those works jointly considers \textit{all} of those modalities.
%
%
\par 
To this end, this work investigates how each of those modalities can benefit the task of multi-person absolute 3D pose and shape estimation.
A particular challenge, however, is that each individual modality has, of course, advantages, but also disadvantages. 
While 2D and 3D keypoint detections can help to infer the local 3D pose of a single person, they cannot ensure 3D consistency across humans and the scene.
Joint angle estimates can be directly used to drive CG characters, but they are usually less accurate than the 3D keypoint detectors due to error accumulation along the kinematic chain.
Normalized disparity maps provide global reasoning of the entire scene as well as the humans in terms of its scale-normalized depth, but they cannot provide absolute depth and scale of the scene.
Finally, human segmentation masks can provide close to pixel-perfect and identity preserving segmentations of humans in the scene, but they lack a 3D understanding.
\par 
Now, to unite all the advantages of each of the modalities while compensating for their potential limitations, \new{we propose the first optimization-based approach that} jointly recovers the absolute 3D position of all humans in the images, their articulated pose, their individual shapes, as well as the scale of the scene from a single video recorded with a static camera; see Fig.~\ref{fig:teaser}.
In particular, we propose a novel energy formulation, which infers the absolute scene depth and the person \finalrev{unique} scale from scale-normalized disparity predictions by using the 2D and joint angle estimates of the humans in the scene as a prior.
Once the per-frame absolute depth is known, we reconstruct a dense point cloud of the static scene in absolute 3D space by segmenting out the humans using the predicted segmentations and aggregating per-frame depth over time.
Finally, we perform a coherent space-time optimization over the entire sequence to ensure temporal and spatial consistency as well as physical plausibility leveraging the aggregated scene estimate and the joint angle predictions.
Note that in each of those steps, the \textit{combination} of different data modalities is leveraged \new{through our method} and only this specific \new{approach} achieves the desired result \new{in the considered setting}, \new{as extensively shown in our results}.
%
%
In summary, our primary technical contributions are as follows:
\begin{itemize}
    \item The first monocular approach for multi-person absolute pose and \finalrev{unique} scale estimation that jointly estimates multiple human poses and the 3D scene by combining data modalities \new{in a novel optimization framework}.
    \item A human body prior to disambiguate the scale of the scene, which allows us to perform a coherent space-time  reasoning of the human motion in absolute space.
    \item We show that the estimated 3D human bodies can be refined in 3D space and time by filtering body movements in 3D coordinates and by penalizing implausible poses w.r.t. the estimated scene, resulting in a more coherent final prediction.
\end{itemize} 
Since our approach estimates joint angles, global positions and scale, the recovered 3D human poses can be directly applied to CG characters enabling exciting applications as shown in Section~\ref{sec:experiments}.
Moreover, we demonstrate that the joint reasoning of the human body shape, pose, and the dense scene over the entire video sequence improves state of the art in terms of 3D localization, scene and person scale, as well as body pose compared to prior work, both, quantitatively and qualitatively. 
Finally, we show that several downstream applications can be directly derived from our method, like monocular human motion capture and avatar control.
%

%
%
\section{Related Work} \label{sec:relatedwork}
3D human motion capture is an active research area, and many works have been proposed in the past~\cite{Mehta2017, Sun_2017_ICCV, martinez2017simple, Chen_2017_CVPR, VIBE2020, zhou2017towards, Consensus_IJCV_2022, TekinKSLF16, Sun_2018_ECCV, Wandt2019}.
Since we target a monocular setting, we do not review multi-view- and depth-based methods.
Instead, we review previous works that are most related to our method. 
%
%
\subsection{3D Human Pose Estimation}
\subsubsection{Single Person Pose Estimation}
Estimating the human body pose in 3D from a single image is a challenging problem that has been successfully handled by learning a human body prior from MoCap data~\cite{ionescu2013human3}.
To simplify the problem, previous methods usually predict 3D coordinates relative to the root joint, assuming a normalized human body size~\cite{Mehta_2017_3DV} and a fixed bounding box around the person in 3D space~\cite{Mehta2017, Pavlakos_2017_CVPR}. 
However, when multiple people are interacting with the environment, normalized and root-relative predictions are not enough to disambiguate the position and scale of individual persons in the scene.
%
\new{%
In addition, directly estimating the 3D joint coordinates could result in implausible poses, which is a problem that can be mitigated by estimating joint angles instead~\cite{zhou2016deep}.
}
%
%
\par
Several works focus on estimating the full human mesh deformation from videos~\cite{xu18, habermann19, habermann20}, assuming that the actor mesh is provided in advance.
Other works for single human estimation~\cite{Kanazawa2018,Kolotouros2019_ICCV,PavlakosChoutas2019} rely on  SMPL~\cite{loper2015smpl} as a proxy shape.
Reconstructing shape proxies along with sparse 3D skeletons is desirable in many scenarios (\eg, they can be used for body parts segmentation).
Moreover, SMPL serves as a statistical prior on human body shapes and enables additional supervisory terms such as human silhouette overlays in 2D, which can result in higher accuracy \cite{PavlakosChoutas2019}. 
%
%
\subsubsection{Multiple Person Pose Estimation} \label{sec:rw_multiperson}
Estimating positions of each person w.r.t. the others is crucial in multi-human pose estimation. 
Nonetheless, most of the existing multi-person methods are by design performing root-relative predictions~\cite{ROMP, benzine2020pandanet, rogez2017lcr, rogez2019lcr}.
Several techniques predict translations of each person in the camera reference frame. 
They either optimize the translation by projecting and fitting the estimated 3D poses into the image plane~\cite{XNect_SIGGRAPH2020, zanfir2021thundr, dabral2019multi} or by directly regressing the distance of the root joint to the camera with a deep neural network~\cite{Moon_2019_ICCV_3DMPPE, lin2020hdnet, wang2020hmor, zhen2020smap}. 
The first case can be more robust to different camera setups, but is limited by the unknown height of each person in the scene.
The second strategy is highly dependent on the training data and may not generalize to camera configurations not present in the training.
\finalrev{Others explore human priors \cite{li2021task} to estimate a global trajectory \cite{yuan2022glamr}, but still fail to recover the body size.}
%
%
\par 
Recent methods performing human depth estimation are focused on penalizing depth ordering of multiple humans. 
For instance, Jiang \etal~\cite{jiang2020coherent} uses instance segmentation masks to penalize depth inversion and Sun \etal~\cite{sun2022putting} proposes to infer the depth of each person based on an imaginary bird's-eye-view representation and to estimate the person age as a proxy for the scale.
Other approaches predict the relative depth among multiple persons by inferring some scene properties. 
A possible scene simplification is to assume a parametric planar floor, in such a way that each prediction can be positioned to respect a plausible human-floor contact~\cite{zanfir2018monocular, ugrinovic2021body}.
The common limitation of such approaches is the dependency on a simplified floor representation, which is often not the case in real applications. \finalrev{Contrarily, we estimate a scene point cloud that can represent a arbitrary ground floor}.
%
%
\par 
The works from Jiang \etal~\cite{jiang2020coherent} and Ugrinovic \etal~\cite{ugrinovic2021body} are the most closely related to ours. 
Similarly to the former, we also render the estimated human models into the image plane to provide additional supervision in the depth dimension, and, related to the latter, we also disambiguate body size and depth for each person by constraining predictions with an estimated scene geometry.
But differently from \cite{jiang2020coherent}, that does not take the scene into account, and from \cite{ugrinovic2021body}, that relies on a simplified scene representation and operates in a single frame, our method represents the scene as a frustum point cloud and performs optimization over the entire video sequence.
In our work, we also rely on a human body proxy model~\cite{loper2015smpl} to estimate joint angles and we propose a new formulation to optimize the position of the humans and the scene in a joint optimization process.
Therefore, our model improves the prediction of human positions by relying on an estimated proxy scene geometry that does not depend on a simplified parametric model.
%
%
\subsection{Scene-aware Motion Capture}
Predicting and understanding how humans interact in 3D has recently gained a lot of attention. 
Several current methods focus on positioning humans in a \finalrev{pre-scanned 3D scene~\cite{HPS, huang2022capturing, PROX_ICCV_2019}} and on simultaneous estimation of human poses and objects humans interact with~\cite{weng2021holistic, chen2019holistic++, yi2022human}.
\finalrev{A different setup assumes an RGB-D sensor \cite{LEMO_zhang2021learning} or a moving camera \cite{zhang2022egobody, liu20214d, henning2022bodyslam, li2022d} that facilitates estimating the scene geometry.}
Recent methods integrate physics-based constraints into monocular 3D human motion capture and mitigate foot-floor penetration and other severe artefacts \cite{PhysCapTOG2020,PhysAwareTOG2021}. 
Yu~\etal~\cite{Yu:2021:MovingCam} also support composite scenes in the parcours and sports scenarios. 
Although there is a growing interest in investigating the interactions of humans and objects~\cite{Dabral_2021_ICCV,bhatnagar2022behave}, 3D motion capture of multiple humans with environmental awareness from a single monocular camera remains underexplored. 
%
%
\par 
Determining the absolute human scale in 3D is an ill-posed and challenging task.
Bieler \etal~\cite{Bieler2019} estimate the height of a single person from monocular videos by observing jumping people. 
Dabral \etal~\cite{Dabral_2021_ICCV} require an interaction with an object undergoing a free flight to resolve the absolute scene scale. 
Both methods assume motion influenced by the universal law of gravity near the surface of Earth, which allows them to relate the time spent in the air or the form of the observed trajectory with absolute distances in the metric units. 
The downside is that jumping humans or flying objects are restrictive assumptions. 
In contrast, we use a human body and 3D scene priors in 3D multi-human motion estimation and do not make strong assumptions about the observed human motions. 
%
%
\begin{figure*}[tbp]
    \centering
    \mbox{} \hfill
    \includegraphics[width=1.0\textwidth]{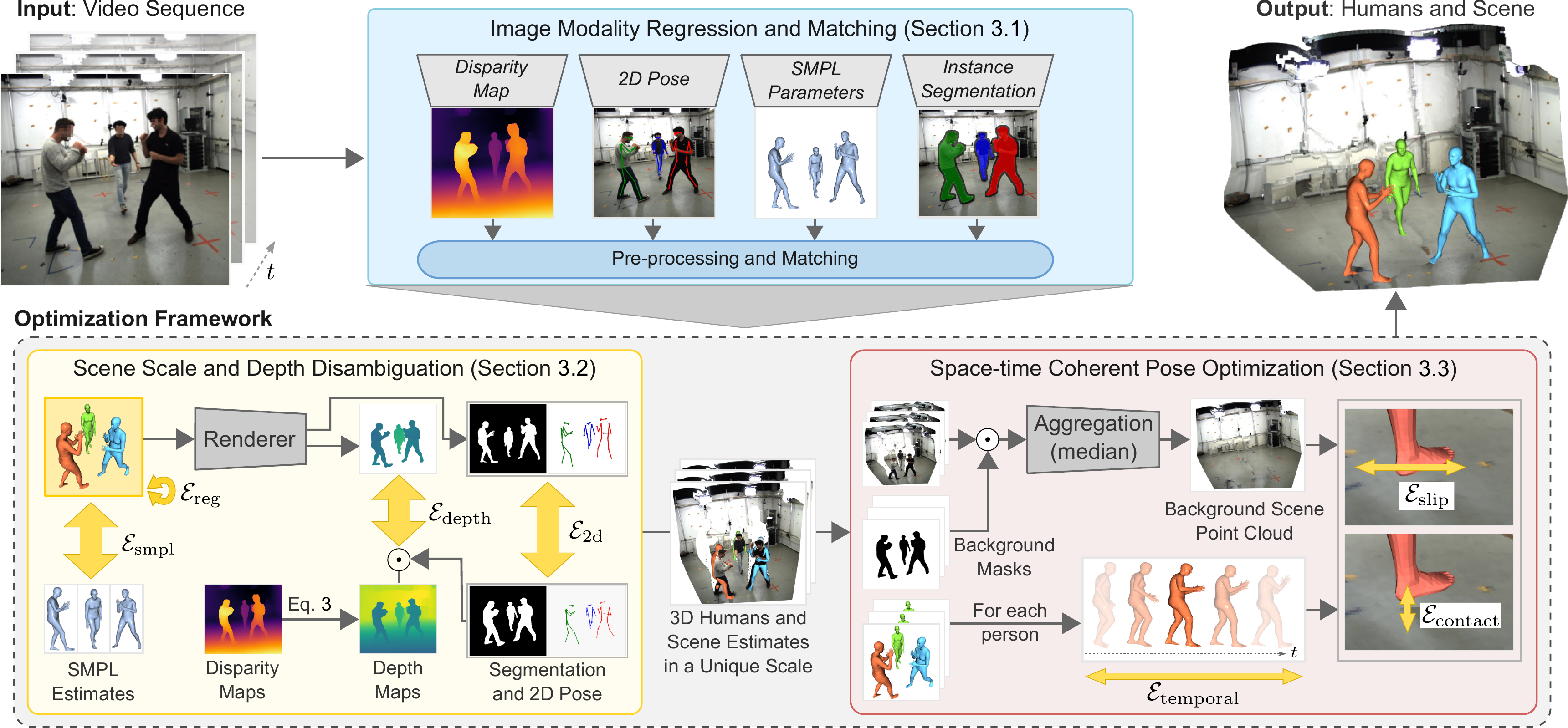}
    \hfill \mbox{}
    \caption{\label{fig:method}%
        \textbf{Overview of our method.}
        For each frame in a monocular RGB video, we first estimate a normalized disparity map, 2D human poses, SMPL model parameters, and segmentation masks.
        These predictions are matched and tracked across frames to obtain per-person associations (blue box). 
        The multi-modal estimates are then fed into our optimization framework. 
        The first part of our optimization process estimates per-frame human models in global position and the scene geometry (yellow box). 
        In the second part, the per-frame scene predictions are aggregated into a single point cloud representation and the human predictions are refined in a space-time coherent manner over the full video (red box).
        The yellow arrows indicate the energy terms minimized by our method.
        The output of our method is the absolute 3D positions of each human in the scene, their shape and pose as well as the scene scale. $\odot$ is the Hadamard product.
    }
\end{figure*}
%
%
\section{Method} \label{sec:method_data}
The goal of our method is to estimate the absolute 3D position of each human in the scene, \finalrev{i.e., up to a unique and global scale}, their proxy shape and pose, as well as the scene scale solely from a monocular RGB video recorded with a static camera for which we know the intrinsics.
To this end, we propose a unified approach that, for the first time, leverages \textit{all} available data modalities, including 2D joint detections, regressed SMPL parameters, estimated disparity maps, and human segmentations.
As illustrated in Figure~\ref{fig:method}, our method is divided into two stages.
\new{%
The first stage, \ie \textit{Image Modality Regression and Matching} (Section~\ref{sec:preprocessing}), extracts per-frame estimates and aggregates human-related predictions to individuals throughout the video sequence.
The second stage, \ie the proposed \textit{Optimization Framework}, estimates the person and \textit{per-frame} scene scale, the \textit{global} 3D position of each person in the scene, as well as the refined articulated body pose in the form of joint angles per frame.}
\par
\new{%
The optimization framework is further subdivided into two parts.
The \textit{Scene Scale and Depth Disambiguation} part (Section~\ref{sec:optim_part1}) recovers a consistent and absolute 3D scene depth per frame, the human scales, and their absolute 3D position and body pose by jointly reasoning about multiple humans and the scene.
The second part, referred to as \textit{Space-time Coherent Pose Optimization} (Section~\ref{sec:optim_part2}),
refines the pose and position of the estimated humans in a space-time coherent formulation, \ie we enforce over the entire sequence the estimated poses to be temporally stable and physically plausible.
For this, we leverage a rough scene geometry estimation, which is obtained by aggregating the absolute depth maps also estimated by our method.
}
This final part significantly reduces artifacts, such as foot sliding, human-scene intersections, and jitter.
Before we explain our method in more detail, we introduce relevant notations.
%
%
\par
\noindent \textbf{Notations.}
The input of our framework is a video sequence $\mathbf{I}_t$, with $t \in \{1, \dots, T\}$, where $T$ is the number of frames. 
We leverage the skinned multi-person linear model (SMPL)~\cite{loper2015smpl} to represent the humans in the scene. 
SMPL is a differentiable parametric human model that takes as input the pose parameters $\boldsymbol{\theta}\in\mathbb{R}^{72}$, corresponding to the axis-angles of $24$ body joints and the global body rotation, and PCA shape parameters $\boldsymbol{\beta}\in\mathbb{R}^{10}$, and produces a skinned human mesh
%
\begin{equation} \label{eq:smpl}
    f_\mathrm{smpl}(\boldsymbol{\theta}, \boldsymbol{\beta}) = \mathbf{V},
\end{equation}
%
where $\mathbf{V}$ are the posed and shaped vertices of the human body; for more details we refer to their paper~\cite{loper2015smpl}.
The mesh vertices regressed by SMPL can also be used to estimate a sparse 3D pose as $\mathcal{J}(\mathbf{V})$, where $\mathcal{J}(\cdot)$ is a linear regressor parameterized by a matrix $\mathbf{W}\in\mathbb{R}^{J\times6890}$, and $J$ denotes the total number of joints. 
\par
To account for translations in 3D space, we further add a translation $\boldsymbol{\Gamma}_{t,n}\in\mathbb{R}^{3}$ to the SMPL representation, \finalrev{where $n$ is the person index}.
Furthermore, the 3D human pose models are overwhelmingly biased towards adult body sizes. 
Thus, we explicitly model the person scale by $s_{n}\in\mathbb{R}^+$ and our final human mesh can be defined as
%
\begin{equation} \label{eq:smplextended}
    \tilde{\mathbf{V}}_{t,n}=s_{n}\mathbf{V}_{t,n} + \boldsymbol{\Gamma}_{t,n}.
\end{equation}
%
This human mesh for person $n$ at time $t$ is then fully determined by the parameters $\boldsymbol{\theta}_{t,n}$, $\boldsymbol{\Gamma}_{t,n}$, $\boldsymbol{\beta}_n$, and $s_{n}$, which we aim to recover in the following.
Important to note is that the person scale $s_{n}$ and shape $\boldsymbol{\beta}_n$ are unique for each person and consistent across the entire video sequence.
%
%
\subsection{Input Modality Regression and Matching} \label{sec:preprocessing}
To solve this underconstrained and challenging problem, our idea is to unite the strength of \textit{all} data modalities, which recent state-of-the-art Computer Vision methods provide, in a single algorithm.
More precisely, we leverage data-driven priors in the form of four off-the-shelf methods for each frame of the input video sequence, as shown in Figure~\ref{fig:method}.
\par
First, we obtain normalized disparity maps $\hat{\mathbf{d}}_t$ from the state-of-the-art DPT model~\cite{ranftl2021vision}, which are then post-processed to enhance sharpness~\cite{watson-2020-stereo-from-mono}.
Note that these maps only encode \textit{relative} and \textit{normalized} depth and they are not consistent across frames, which becomes visible in the form of depth jitter.
\par
Second, 2D pose tracking is obtained by AlphaPose~\cite{fang2017rmpe}, which coherently \finalrev{detects and} tracks 2D joint positions $\hat{\mathbf{P}}^\mathrm{2d}_{t,n}\in\mathbb{R}^{J\times{2}}$ in image space and over time.
Although this method is very robust due to training on large scale data, it falls short in predicting 3D.
\par
Third, we predict the body shape $\boldsymbol{\beta}_{t,n}$ and joint angles $\hat{\boldsymbol{\theta}}_{t,n}$ for each person in each frame using ROMP~\cite{ROMP}. 
Since ROMP predicts varying shapes for a single person across time, we average the predictions over the entire sequence to obtain a temporally consistent body shape.
Thus, the vertices (Equation~\ref{eq:smplextended}) are now only a function of the pose $\boldsymbol{\theta}_{t,n}$, translation $\boldsymbol{\Gamma}_{t,n}$, and scale $s_{n}$, which will be important in the next section.
Moreover, to match the 2D AlphaPose and the SMPL detections, we leverage ROMPs projection model, compute the average Euclidean distance in image space, and pair detections with the lowest distance based on the Hungarian matching.
It is worth mentioning that ROMP cannot account for out-of-distribution body sizes, e.g. small kids, neither it can predict the absolute 3D position of the humans with respect to the scene.
\par
Fourth, we also leverage human segmentation masks, referred to as $\boldsymbol{\Omega}_{t,n}\in\mathbb{R}^{H\times{W}}$, which are obtained from Mask2Former~\cite{cheng2021masked}. 
Similarly, if we consider all the remaining pixels for frame $t$ that do not belong to a person mask, we can also obtain a per-frame background segmentation mask $\mathbf{B}_t\in\mathbb{R}^{H\times{W}}$.
To ensure that the 2D AlphaPose detections, the SMPL detections, and the foreground masks have a consistent person ID, we read the pixel values of the segmented masks at the 2D joint detections for each detected skeleton and apply a max-voting to retrieve the ID of the person.
\par
In summary, the inputs to our algorithm now are:
\begin{itemize}
    \item{$\hat{\mathbf{d}}_t$: Normalized disparity maps}
    \item{$\hat{\mathbf{P}}^\mathrm{2d}_{t,n}$: 2D joint predictions}
    \item{$\hat{\boldsymbol{\theta}}_{t,n},\hat{\boldsymbol{\beta}}_{n}$: Pose angle and shape estimates}
    \item{$\boldsymbol{\Omega}_{t,n},\mathbf{B}_t$: Human and background segmentations}
\end{itemize}
Note that none of these predictions individually or by a trivial combination is discriminative enough to fully describe the entire scene, \ie absolute 3D position, pose, and scale of the humans in the scene.
Next, we demonstrate how our proposed method solves this problem.
%
%
%
%
%
\subsection{Scene Scale and Depth Disambiguation} \label{sec:optim_part1}
%
%
\new{In the first part or our optimization process} we focus on jointly obtaining the joint angles $\boldsymbol{\theta}_{t,n}$, shape parameters $\boldsymbol{\beta}_{n}$, global translation $\boldsymbol{\Gamma}_{t,n}$, and scale $s_{n}$ of each person.
Importantly, this step is performed jointly for the entire sequence, \finalrev{where the global reference is in the static camera}. 
However, estimating the height of a person \textit{and} the scale given only a single RGB video is, by itself, an ill-posed problem as variations in scale can be compensated by a translation along the depth and vice versa.
As a result, infinitely many scale/translation combinations can lead to the same 2D image projections.
\par 
So far, we only considered individual humans without looking at the surrounding scene, although the scene itself can provide an important prior that helps to solve the above problem.
Therefore, we leverage recent advances in monocular depth estimation~\cite{ranftl2021vision}, which regress per-pixel normalized disparity maps $\hat{\mathbf{d}}_t$. 
It encodes the relative depth of each person in the scene, but obtaining the absolute depth values solely from $\hat{\mathbf{d}}_t$ is also an ill-posed problem, and further these predictions are not consistent across frames.
The question remains, how the absolute scene depth or equivalently the human scales and translations can be recovered.
\par
Our idea is to set the two entities, i.e., the scene and the humans, into a relation such that they constrain each other in an absolute 3D space.
While the humans can already be represented in absolute space by means of their global translation $\boldsymbol{\Gamma}_{t,n}$ and scale $s_{n}$, we also require a per-frame conversion of temporally inconsistent normalized disparity maps to absolute depth maps, which can be defined as
%
\begin{equation}
    \tilde{\mathbf{D}}_t=\frac{%
            z_{\mathrm{far},t} z_{\mathrm{near},t}%
        }{%
            \hat{\mathbf{d}}_t(z_{\mathrm{far},t}-z_{\mathrm{near},t})+z_{\mathrm{near},t}%
        }
    \label{eq:inv_depth}
\end{equation}
%
where $z_{\mathrm{near},t}$ and $z_{\mathrm{far},t}$ are the near and far depth values, respectively. 
Intuitively, this operation shifts and scales the normalized disparity maps to convert them to absolute depth values.
Importantly, these near and far values are optimized per-frame to compensate for the temporal inconsistencies in the disparity maps.
\par
Once both humans and the scene can be represented in absolute 3D space, we now relate them to each other by jointly solving for $\boldsymbol{\kappa}_{t,n} \in \{z_{\mathrm{near},t}, z_{\mathrm{far},t}, \boldsymbol{\theta}_{t,n}, \boldsymbol{\beta}_{n}, \boldsymbol{\Gamma}_{t,n}, s_{n}\}$ by minimizing the energy
%
\begin{equation}
    \argmin_{\forall t \in \{1,...,T\}, \forall n \in \{1,...,N\}: \boldsymbol{\kappa}_{t,n}}    E_\mathrm{I}, \quad \text{with} 
    \label{eq:optim_part1}
\end{equation}
\begin{equation}
    E_\mathrm{I} = \mathcal{E}_\mathrm{depth}+\mathcal{E}_\mathrm{2d}+\mathcal{E}_\mathrm{smpl}+\mathcal{E}_\mathrm{reg},
    \label{eq:energy_part1}
\end{equation}
%
which is jointly optimized over the entire sequence.
In particular, our energy is composed of a depth term $\mathcal{E}_\mathrm{depth}$, a 2D image evidence term $\mathcal{E}_\mathrm{2d}$, a joint angle and shape term $\mathcal{E}_\mathrm{smpl}$, and additional regularization terms $\mathcal{E}_\mathrm{reg}$. 
In the following, we explain each term in more detail. 
%
%
\subsubsection{Depth Consistency Energy}
Most importantly, to ensure a coherent depth between the scene and all humans in the scene, we propose a depth consistency energy
%
\begin{equation}
    \mathcal{E}_\mathrm{depth}=\lambda_\mathrm{depth}\sum_{t,n}%
      {\left({%
      M(\Psi_{d}(\tilde{\mathbf{V}}_{t,n}))-%
      M(\tilde{\mathbf{D}}_t)}%
      \right)^2},
    \label{eq:energydepth}
\end{equation}
\begin{equation}
    M(\mathbf{D})=\sum\frac{\boldsymbol{\Omega}_{t,n}}{\lvert \boldsymbol{\Omega}_{t,n} \rvert}\log(\mathbf{D}),
\end{equation}
%
%
where $\lvert \boldsymbol{\Omega} \rvert$ denotes the number of foreground pixels, \new{$M(\cdot)$ computes the average of the log-depth in the foreground}, $\Psi_{d}(\cdot)$ is a differentiable rasterizer~\cite{ravi2020pytorch3d} that \finalrev{projects and} converts a 3D mesh into a depth map in the image plane, and $\lambda_\mathrm{depth}$ is a \new{hyperparameter}.
The vertices $\tilde{\mathbf{V}}_{t,n}$ refer to the estimated SMPL models of each person in global space, which are a function of the variables $\boldsymbol{\theta}_{t,n}$, $\boldsymbol{\beta}_{n}$, $\boldsymbol{\Gamma}_{t,n}$, and $s_{n}$ (Equation~\ref{eq:smplextended}). 
The rasterized human depths are then compared to the estimated absolute depth map $\tilde{\mathbf{D}}_t$ of the scene (Equation~\ref{eq:inv_depth}), which are a function of the variables $z_{\mathrm{near},t}$ and $z_{\mathrm{far},t}$. 
Thus, this energy jointly optimizes the human \textit{and} the scene parameters.
However, since both sides of the penalty term contain free variables, this energy alone would not disambiguate the problem.
%
%
\subsubsection{Image Projection Energy}
We introduce an additional data term, which further constrains the human-related variables \new{by enforcing the 3D bodies to project accurately into the image plane. More precisely, the data term}
\begin{equation}
\label{eq:2d}
    \mathcal{E}_\mathrm{2d}=\mathcal{E}_\mathrm{joints}+\mathcal{E}_\mathrm{silhouette}
\end{equation}
penalizes the error between the projected 3D body joints $\mathcal{J}(\tilde{\mathbf{V}}_{t,n})$ of the optimized SMPL models and the respective 2D body joints $\hat{\mathbf{P}}^\mathrm{2d}_{t,n}$ regressed by AlphaPose with
\begin{equation}
\label{eq:joints}
    \mathcal{E}_\mathrm{joints}=%
    \sum_{t,n}{%
        \left\Vert{%
            \Pi(\mathcal{J}(\tilde{\mathbf{V}}_{t,n}))%
            - \hat{\mathbf{P}}^\mathrm{2d}_{t,n}%
        }\right\Vert_2^2},
\end{equation}
where $\Pi(\cdot)$ is the perspective camera projection operator.  
\new{%
The right term of \eqref{eq:2d} penalizes the discrepancy between the SMPL silhouette and the instance segmentation masks: 
}
\begin{equation}
\label{eq:silhouette}
    \mathcal{E}_\mathrm{silhouette}=%
    \frac{\lambda_{silhouette}}{\lvert \boldsymbol{\Omega} \rvert}\sum_{t,n}{%
        \boldsymbol{\sigma}_{t,n}%
        \left\Vert{%
            \Psi_{s}(\tilde{\mathbf{V}}_{t,n})%
            - \boldsymbol{\Omega}_{t,n}%
        }\right\Vert_2^2},
\end{equation}
\new{%
where $\Psi_{s}(\cdot)$ is a differentiable renderer~\cite{ravi2020pytorch3d} that \finalrev{projects and} converts a 3D mesh into a silhouette image and $\boldsymbol{\sigma}_{t,n}$ is a visibility mask, so vertices hidden by other humans are not penalized.
}
%
%
\subsubsection{Joint Angle and Shape Energy}
Since \eqref{eq:2d} only constrains the parameters in 2D image space, we further add an additional data term that ensures that the optimized SMPL parameters are close the prediction of ROMP: 
%
\begin{equation}
    \mathcal{E}_\mathrm{smpl}=%
    \lambda_\mathrm{smpl}%
    \sum_{t,n}
        \left\Vert \boldsymbol{\theta}_{t,n}%
        - \hat{\boldsymbol{\theta}}_{t,n}
        \right\Vert_1%
        +%
        \left\Vert \boldsymbol{\beta}_{n}%
        - \hat{\boldsymbol{\beta}}_{n}
        \right\Vert_1.%
\end{equation}
%
Here, $\left\Vert \cdot \right\Vert_1$ denotes the $L1$ norm.
%
%
\subsubsection{Temporal and Human Priors}
To further constrain the scale and position of a person, we leverage priors on the human body size and on the temporal information. 
This is achieved by our regularization term
%
\begin{equation}
    \mathcal{E}_\mathrm{reg}=\mathcal{E}_\mathrm{scale}+\mathcal{E}_\mathrm{speed}.
\end{equation}
%
\par 
For the scale term $\mathcal{E}_\mathrm{scale}$, our assumptions are two-fold: 
\textit{i)} The scale of a person should not deviate too much from the standard person size, \finalrev{i.e., the standard SMPL size when $s_n=1$}, and \textit{ii)} the average scale of multiple people in the scene should remain close to one. 
This dual assumption is enforced by 
%
\begin{equation}
    \mathcal{E}_\mathrm{scale}=\lambda_\mathrm{scale}\sum_{n}\left(s_n - 1\right)^2
    + \left(\sum_{n}(s_n - 1)\right)^2,
\end{equation}
\finalrev{where the first term accounts for the individual person scale and the second term accounts for the average scale of multiple persons}.
%
%
\par 
In addition to the person scale, we also introduce an underlying assumption that locomotion is rather smooth over time based on the physical limits of the human body, so we penalize large movements of the root joint by our energy
%
\begin{equation}
    \mathcal{E}_\mathrm{speed}=\lambda_\mathrm{speed}\sum_{t,n}\left\Vert \boldsymbol{\Gamma}_{t,n}-\boldsymbol{\Gamma}_{t-1,n}\right\Vert_2^2.
\end{equation}
%
\par
\begin{figure}
    \centering
    \includegraphics[width=0.474\textwidth]{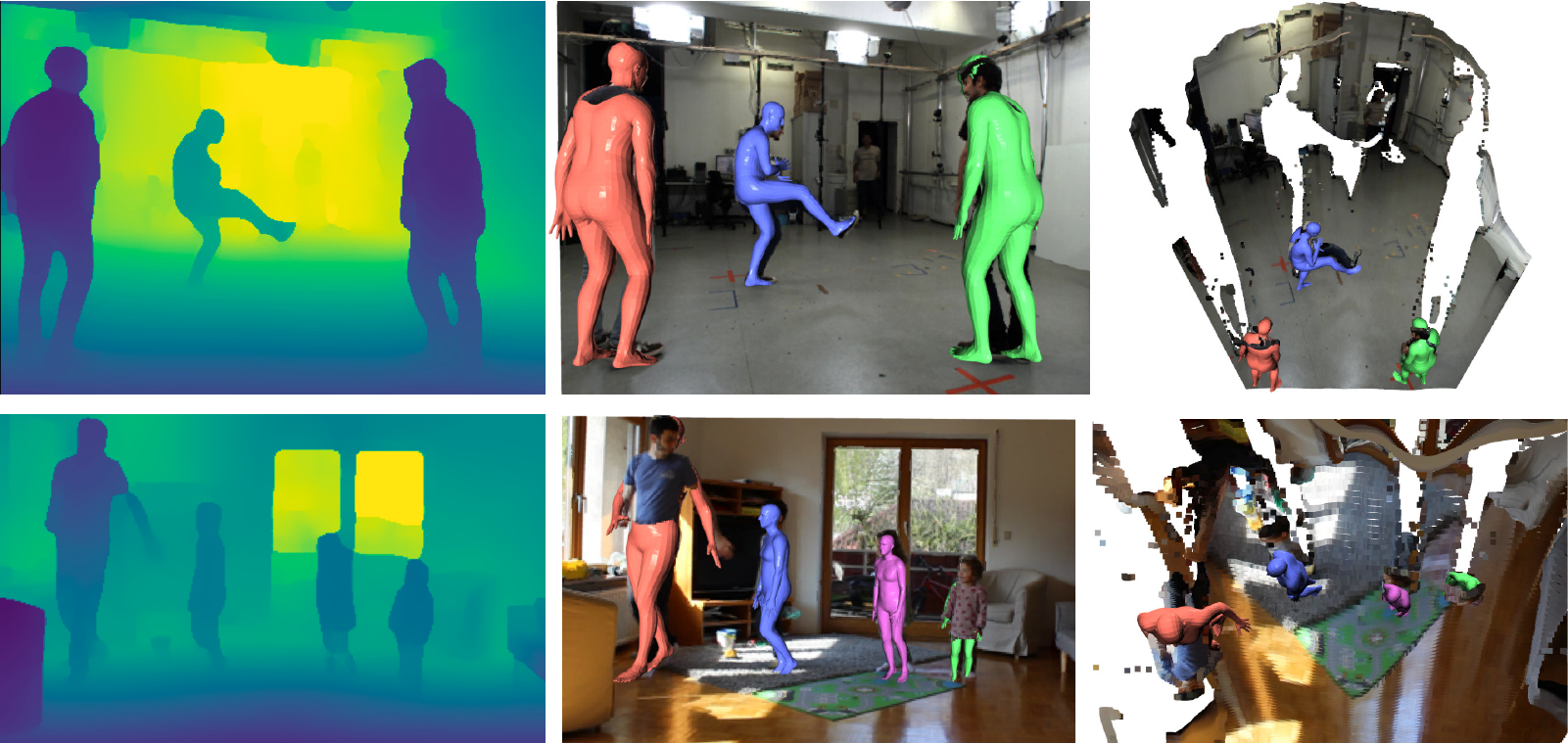}
    \caption{\label{fig:qualitative_stage1}\new{%
        Per-frame estimations of our method considering the first optimization part (Section~\ref{sec:optim_part1}) only. From left to right: Estimated depth map, frontal view of the scene and estimated humans, and top view. 
        Note how the persons' absolute 3D location, articulated pose and shape as well as the scene scale can be recovered from a single input image, even with people of different sizes (bottom row).
        }
    }
\end{figure}

 
\new{In the optimization process described above,} the \textit{per frame} human parameters and the absolute scene depth are obtained by means of the optimized human $\forall{t}\in\{1,...,T\},\forall{n}\in\{1,...,N\}: \boldsymbol{\theta}_{t,n}, \boldsymbol{\beta}_{n}, \boldsymbol{\Gamma}_{t,n}, s_{n}$,  and scene $z_{\mathrm{near},t}, z_{\mathrm{far},t}$ parameters.
Figure~\ref{fig:qualitative_stage1} shows our estimated scene and humans for example frames.
Note that the estimated depth looks plausible, humans and the scene are coherent with each other, and the reprojection of humans into the input view looks accurate.
%
%
%
\subsection{Space-time Coherent Pose Optimisation} \label{sec:optim_part2}
\new{%
Since we obtained absolute and per-frame human models and scene estimations, both information can be used together to further refine the human poses in a spatially and temporally coherent manner. Therefore, in the last part of our optimization method, we refine the estimated poses in 3D by enforcing physical plausibility between humans and the estimated scene, as well as by applying a temporal smoothness term.
}
%
%
More precisely,
\new{%
we extend \eqref{eq:optim_part1} by including a new energy term $E_\mathrm{II}$:
}
%
\begin{equation}
    \argmin_{\forall t \in \{1,...,T\}, \forall n \in \{1,...,N\}: \boldsymbol{\kappa}_{t,n}}    E_\mathrm{I}+E_\mathrm{II}, \quad \text{with} 
    \label{eq:optim_part2}
\end{equation}
\begin{equation}
    E_\mathrm{II} = \mathcal{E}_\mathrm{contact}+\mathcal{E}_\mathrm{slip}+\mathcal{E}_\mathrm{temporal}.
    \label{eq:energy_part2}
\end{equation}
%
%
%
\new{%
For implementing $\mathcal{E}_\mathrm{contact}$ and $\mathcal{E}_\mathrm{slip}$, we leverage the estimated scene geometry as a reference for enforcing foot contact and penalizing foot slipping.
In the following, we first explain how the per-frame depth maps are aggregated into a static 3D scene representation, then we present the energy terms of $E_\mathrm{II}$ in more detail.
}
%
%
\subsubsection{Scene Point Cloud Estimation}
%
%
\new{%
Our method relies on humans as anchors in the scene, \ie, the estimated geometry around the humans tends to be coherent. However, mainly due to occlusions, the estimated per-frame absolute depth values are not yet temporally consistent for the whole scene. To obtain a static representation of the background, we rely on the segmentation masks to aggregate the depth values in the background from each frame into a single depth map. This static depth map representation is obtained by computing the per-pixel \textit{median} for the entire video sequence, which is a metric robust to outlier depth values. We also experimented with more sophisticated aggregation strategies, such as aggregating values near the human anchors weighted by a Gaussian distribution---since the human positions are stable---but this strategy was significantly more expensive and resulted in marginal improvements.
}
At the end of this aggregation process, we obtain a single depth map $\hat{\mathbf{D}}$ of the scene, which can be then converted to a point cloud representation $\mathbf{P}\in\mathbb{R}^{H W\times{3}}$ in absolute 3D space.
%
%
\subsubsection{Improving Physical Plausibility of Estimated Motions}
Recently, a series of works highlighted the importance of physics awareness in monocular \textit{single} person motion capture~\cite{PhysCapTOG2020,PhysAwareTOG2021,rempe2021humor,li2022d} \new{with assumptions about the camera and floor plane positions}.
\new{%
Inspired by them and the fact that we obtain a coherent and \finalrev{unique} scale estimation of the scene, we propose to model in our energy formulation the physical interaction between the humans and the environment.
}
Here, the first term penalizes "floating" characters, \ie, humans that are not in contact with the ground, and the second term penalizes foot sliding, \ie, a foot that is in contact with the ground should not move.
\par
More precisely, given the scene point cloud $\mathbf{P}$ and the estimated human meshes $\tilde{\mathbf{V}}_{t,n}$, floating characters are penalized by
%
\begin{equation}
    \mathcal{E}_\mathrm{contact}=\lambda_\mathrm{contact}\sum_{t,n}\zeta\left(\left\Vert\text{min}(\tilde{\mathbf{V}}^{y+}_{t,n}-\mathbf{P})\right\Vert_1\right)
\end{equation}
%
where $\tilde{\mathbf{V}}^{y+}_{t,n}\in\mathbb{R}^{1\times{3}}$ is the vertex of person $n$ at time $t$ with lower $Y$ coordinate, considering that the $Y$-axis is the gravitational axis for our coordinate frame.
In other words, the term $\mathcal{E}_\mathrm{contact}$ minimizes the distance between the lower vertex $\tilde{\mathbf{V}}^{y+}$ of each prediction and its respective closest point in the scene point cloud.
Here, $\zeta(\cdot)$ is a robust thresholding function, which only considers distances below $20cm$.
\par 
The term
%
\begin{equation}
    \mathcal{E}_\mathrm{slip}=\lambda_\mathrm{slip}\sum_{t,n}{\zeta}\left\Vert\Delta(\tilde{V}^{y+}_{t,n})\right\Vert_1
\end{equation}
%
penalizes the movement of this lowest vertex in the time domain ($\Delta$) when it is in contact with the scene.
%
%
%
By applying those energy terms, we can now enforce that the humans interact more physically accurate with respect to the 3D scene.
%
%
\subsubsection{Temporally Stable Pose}
Furthermore, since the joint and absolute position optimized by $E_\mathrm{I}$ can still contain smaller jitter, we propose a temporal stability term
%
\begin{equation}
    \mathcal{E}_\mathrm{temporal}=\lambda_\mathrm{temporal}\sum_{t,n}\left\Vert\Delta_t({\tilde{\mathbf{V}}}_{t,n})-\Delta_t({\bar{\mathbf{V}}}_{t,n})\right\Vert^2,
    \label{eq:temporal}
\end{equation}
%
based on the 1\EUR{} filter~\cite{one_euro_filter},
where $\Delta_t(\mathbf{V}_{t,n})=\mathbf{V}_{t,n}-\mathbf{V}_{t-1,n}$ is the temporal variation of the human mesh vertices and $\bar{\mathbf{V}}_{t,n}$ are the estimated SMPL vertices after temporal filtering~\cite{one_euro_filter}.
This term allows us to obtain temporally more stable poses
\new{%
with significantly less jitter.
}
%
%

\section{Experiments} \label{sec:experiments}
In this section, we present an empirical evaluation of our method. We first briefly describe the datasets and metrics used in our experiments in Sections~\ref{sec:datasets} and \ref{sec:metrics}, followed by the implementation details in Section \ref{sec:details}. Next, we compare our approach with the most related works to ours in Section~\ref{sec:comparison_sota}. In Section~\ref{sec:ablation}, we perform a thorough ablation study of the main components of our method and show additional qualitative results in  Section~\ref{sec:qualitative}. 
\subsection{Datasets} \label{sec:datasets}
\textbf{MuPoTs-3D}~\cite{singleshotmultiperson2018} is a test dataset composed of $20$ video sequences with multiple people, including different types of cameras in indoor and outdoor environments. We followed the evaluation protocol from \cite{singleshotmultiperson2018} in our experiments. This dataset is especially challenging due to the large amount of interactions between humans and the various types of scenes. Ground-truth 3D pose annotations are provided in absolute coordinates. 
\par
\noindent
\new{%
\textbf{CMU Panoptic}~\cite{PanopticData} is a dataset recorded in the Panoptic Studio with multiple people. As in preliminary work~\cite{jiang2020coherent, zanfir2018monocular}, we use this dataset for evaluation considering the sequences \texttt{haggling1}, \texttt{ultimatum1}, and \texttt{pizza1}, which are performed by several adults.
}
\par
\new{%
In addition to the previous datasets, we also evaluated our method quantitatively on Internet videos considering challenging cases with multiple people of different sizes, including adults and children.
}
\subsection{Metrics} \label{sec:metrics}
\finalrev{\textbf{MRPE and AP}}. We quantitatively evaluate the prediction of the absolute 3D location of a human using the widely adopted mean root position error (MRPE), in millimeters, and the average precision of the human root joint (AP${}_{25}^{root}$) \cite{Moon_2019_ICCV_3DMPPE}, considering the standard threshold of $25$ cm.
\par
\noindent
\finalrev{\textbf{3DPCK}}. The quality of the articulated 3D pose prediction is measured using root-relative 3DPCK~\cite{Mehta_2017_3DV}, with the standard threshold of $15$ cm. 
The 3DPCK metric enables measuring the correctness of the pose, independently of the prediction of the absolute 3D location of the human.

\noindent
\finalrev{\textbf{MPJPE}}. \new{%
For a fair comparison with previous methods, we also report  root-relative mean per-joint position error (MPJPE) in the CMU Panoptic dataset.
}
\par
\noindent
\finalrev{\textbf{Jitter}}. Finally, since we are targeting high-quality temporal predictions in 3D coordinates, we also evaluate the amount of jitter of our estimations, which is a critical indicator for many downstream applications. For this evaluation, we adapted the \textit{temporal smoothness error} $e_\mathrm{smooth}$ from \cite{PhysAwareTOG2021} to evaluate the jitter in 3D coordinates.
\subsection{Implementation Details} \label{sec:details}
Our method is implemented in PyTorch~\cite{NEURIPS2019_9015} using PyTorch3D~\cite{ravi2020pytorch3d} for the rasterization \eqref{eq:energydepth} and silhouette rendering \eqref{eq:silhouette}.
\finalrev{The camera intrinsics are used in the 3D joint projection \eqref{eq:joints}, rasterization \eqref{eq:energydepth}, and rendering \eqref{eq:silhouette} parts, and can be obtained from video metadata if not given}.
We apply the RMSprop~\cite{hinton2012neural} optimizer with the parameters $\alpha$ and \textit{momentum} set to $0.5$ and $0.9$, respectively, for all experiments.
\new{%
In the optimization process, we initially minimize the first part \eqref{eq:optim_part1} only for $30$ iterations, then perform the full optimization \eqref{eq:optim_part2} for more $200$ iterations.
We use a learning rate initially set to $0.01$ and exponentially decaying with factor $0.99$.
}
\new{%
The weights $\lambda_{(.)}$ were empirically defined to balance the magnitude of the individual energy terms, and fixed in the method in all experiments, except when mentioned otherwise (ablation in Section~\ref{sec:ablation}).
}
\new{%
The values were defined as $\lambda_\mathrm{depth}=\lambda_\mathrm{speed}=0.05$, $\lambda_{silhouette}=0.1$, 
$\lambda_\mathrm{smpl}=\lambda_\mathrm{temporal}=0.002$, $\lambda_\mathrm{scale}=0.0001$, $\lambda_\mathrm{contact}=0.001$, and $\lambda_\mathrm{slip}=0.01$.
}
For numerical stability, we constrain the variables $s_{n}$, $z_{\mathrm{near},t}$, and $z_{\mathrm{far},t}$ to be non-zero and positive.
Both human and background segmentation masks were post-processed with morphological erosion and dilation filters of size $3{\times}3$ and $5{\times}5$, respectively.
For the sake of GPU memory efficiency, we use mini batches of ten images in the depth and silhouette losses. Our experiments run on a workstation with one Nvidia Titan V GPU with 12 GB of memory.

\begin{table}[]
\caption{\label{tab:comparison_sota_ap_pck}%
Comparison of our method with previous approaches on MuPoTs-3D 
in the MRPE (lower is better), AP${}_{25}^{root}$, and 3DPCK metrics (higher is better), considering the global 3D pose and the normalized (univ) ground truths. 
Our approach is superior to all compared methods on the absolute metrics (MRPE, AP${}_{25}^{root}$ and  3DPCK${}_{3d}$), \ie, the most expressive ones for 3D human motion capture. 
{}``$^\dagger$'' evaluated on samples with IK only; {}``$^{*}$'' evaluated on root-relative predictions without IK; {}``$^\ddag$'' results only possible with an additional 2D fitting stage, implemented as our baseline.
}
\begin{adjustbox}{width=0.48\textwidth,center}
\begin{tabular}{@{}lccccc@{}}
\toprule
Method & \begin{tabular}[c]{@{}c@{}}Char.\\ control\end{tabular} & MRPE~$\downarrow$ & AP${}_{25}^{root}$ & 3DPCK${}_{3d}$ & 3DPCK${}_{univ}$ \\ \midrule
LCR-Net~\cite{rogez2017lcr}  & \cxmark & -- & --  &  -- & 53.8 \\
LCR-Net++~\cite{rogez2019lcr}  & \cxmark & -- & --  &  -- & 70.6 \\ 
3DMPPE~\cite{Moon_2019_ICCV_3DMPPE}  & \cxmark & -- & 31.0  & -- & \textbf{81.8} \\
SMAP~\cite{zhen2020smap} & \cxmark & -- & 45.5  & -- & 80.3 \\
XNect$^{*}$~\cite{XNect_SIGGRAPH2020}  & \cxmark & -- & --  &  64.1 & 71.9 \\
XNect$^\dagger$~\cite{XNect_SIGGRAPH2020}  & \ccmark  & 639 & 31.6 &  56.5  & 60.1 \\
CRMH~\cite{jiang2020coherent} & \ccmark & -- & -- & -- & 69.1 \\
BEV~\cite{sun2022putting} & \ccmark & -- & -- & -- & 70.2 \\ \midrule
Baseline (ROMP$+$2D fitting)   & \ccmark  & 331$^\ddag$ & 45.4$^\ddag$  &  68.2$^\ddag$ & 71.8 \\
\textbf{Ours} & \ccmark & \textbf{266} & \textbf{62.3} & \textbf{74.9} & \underline{78.9} \\ \bottomrule 
\end{tabular}
\end{adjustbox}
\end{table}

\subsection{Comparison with Previous Methods} \label{sec:comparison_sota}

In Table~\ref{tab:comparison_sota_ap_pck}, we compare our method to the most related prior work.
\new{%
We compare our method for human localization considering MRPE and AP${}_{25}^{root}$ metrics with the methods that are capable of providing such predictions.}
We use two protocols to evaluate the quality of the 3D pose. First, we compare against the global 3D pose without any normalization, which is a fairer protocol for our method, since we are capable of estimating the person scale (denoted by 3DPCK${}_{3d}$). In the second case, we compare against the \textit{universal} 3D pose, which has all bone lengths normalized to a standard size, as described in \cite{singleshotmultiperson2018} (denoted as 3DPCK${}_{univ}$). For this universal protocol, in our method, we assume person scale $s_{n}$ equals to one for all predictions.
\new{%
Note how our method outperforms all prior work by a wide margin at 3D localization and also performs better at estimating the articulated pose compared to all other methods that allow for character control.
}
\new{%
As a baseline, we evaluate ROMP~\cite{ROMP} predictions with an additional stage for fitting estimated SMPL models to AlphaPose 2D body joint detections, \finalrev{since this is the closest setup to our method without including our new energy functions}. For this, we assume a unitary person scale (w.r.t. the SMPL neutral model) and optimize only the global translation in 3D of each person.
}
In a similar manner, XNect~\cite{XNect_SIGGRAPH2020} estimates the global position by fitting the predicted 3D poses into 2D body joints, assuming a universal and normalized human body size. The inverse kinematics (IK) stage from XNect allows this global estimation, however, since the optimized 3D human pose differs from the preliminary estimated pose, the accuracy after IK drops significantly.
In summary, we observe that our approach outperforms previous methods for human position estimation by a significant margin, improving the average precision of the root joint from $45.4\%$ to $62.3$\%. Our method also outperforms all other approaches for human pose estimation that are capable of driving a virtual character. 
%
%
\begin{table}[]
\caption{\label{tab:comparison_panoptic_stairs}%
Comparison of our method with previous approaches on the CMU Panoptic dataset for 3D pose estimation. Results reported in millimeters. Camera views capturing only the upper body parts were not used in our evaluation. \new{${}^\dagger$ evaluated in all the sequences.
\finalrev{Best results are \textbf{bold} on the standard sequences and \underline{underlined} on the full-body visible sequences.}}
}
\begin{adjustbox}{width=0.47\textwidth}
\begin{tabular}{@{}llcccc@{}}
\toprule
Metric & Method & Haggling & Ultimatum & Pizza & Avg. \\ \midrule
\multirow{3}{*}{MPJPE} & CRMH~\cite{jiang2020coherent}$^\dagger$ & 129.6 & 153.0 & 156.7 & 146.4 \\
                       & BEV~\cite{sun2022putting}$^\dagger$ & \textbf{90.7} & \textbf{113.1} & \textbf{125.2} & \textbf{109.6} \\ \cline{2-6} 
                       & Baseline & 93.6  & 133.8 & 145.9 & 124.4 \\ \cline{2-6} 
                       & \underline{}{Ours} & \underline{84.5}  & \underline{108.9} & \underline{133.2} & \underline{108.9} \\ \midrule
\multirow{2}{*}{MRPE}  
                       & Baseline      & 235.2 & 269.6 & 356.4 & 287.0 \\ \cline{2-6} 
                       & \underline{Ours} & \underline{213.7} & \underline{208.0}  & \underline{229.7} & \underline{217.1} \\ \bottomrule
\end{tabular}
\end{adjustbox}
\end{table}
%
%
\par
\new{%
In Table~\ref{tab:comparison_panoptic_stairs}, we compare our method with other approaches on the CMU Panoptic dataset. This dataset is specially challenging because in many sequences the persons are only partially visible, either due to occlusions, or because the camera is capturing only the upper body part of the actors. Even in this challenging scenario, our method performs on par with the recent BEV~\cite{sun2022putting} method, which was trained on the CMU Panoptic dataset and, therefore, performs better in the cases of partial body visibility then our optimization approach. In order to evaluate the performance of our method on the more practical scenario of cameras recording the full body of the persons, we removed the few sequences capturing only the upper body parts. In this setup, we largely improve over other methods and over our baseline, as can be seen by the underlined numbers in Table~\ref{tab:comparison_panoptic_stairs}.
}
\par
For many downstream applications, such as gaming and character control, jitter is a severe artifact that hinders usability. Therefore, we also evaluated our method by reporting the temporal smoothness error $e_\mathrm{smooth}$ in 3D coordinates. The results from our method, as well as from previous work in the literature \new{related to ours}, are shown in Table~\ref{tab:comparison_sota_jitter}.
In this experiment, we compared our approach with two methods from the literature, showing a significant improvement in reducing the jitter artifact.
\new{%
Furthermore, we also evaluated the contribution of different components of our method. For instance, the temporal energy term in our approach has a critical effect in reducing jitter. In addition, the contact and slip terms also contribute in a small proportion but consistently to all metrics, regardless the presence or absence of the temporal energy.
}
When all terms are included, our approach is very stable, with an average jitter error below $1cm$.
%
\begin{table}[]
\caption{\label{tab:comparison_sota_jitter}%
Comparison of our method on MuPoTs-3D with previous approaches on temporal smoothness error $e_\mathrm{smooth}$, that measures the amount of jitter in the predictions in millimeters.
We also report the MRPE and 3DPCK${}_{3d}$ metrics for completeness.
Our method has a drastically lower jitter in the prediction compared to previous multi-person motion capture approaches.
}
\begin{adjustbox}{width=0.48\textwidth,center}
\begin{tabular}{@{}lccc@{}}
\toprule
Method & \multicolumn{1}{l}{Jitter~$\downarrow$} & MRPE~$\downarrow$ & 3DPCK${}_{3d}$~$\uparrow$ \\ \midrule
XNect~\cite{XNect_SIGGRAPH2020}  & 136.4 & 639 & 56.5 \\
ROMP~\cite{ROMP}                 &  59.6 & 331 & 68.2 \\ \midrule
Ours ($E_\mathrm{I}$ only)                  &  17.5  & 281 & 73.5 \\ 
Ours ($E_\mathrm{I}+\mathcal{E}_\mathrm{contact}$)   &  17.6  & 276 & 73.7 \\ 
Ours ($E_\mathrm{I}+\mathcal{E}_\mathrm{contact}+\mathcal{E}_\mathrm{slip}$) &  17.1  & 273 & 73.8 \\ 
Ours ($E_\mathrm{I}+\mathcal{E}_\mathrm{temporal}$) & 7.8  & 272  & 74.8 \\ 
\textbf{Ours} ($E_\mathrm{I}+\mathcal{E}_\mathrm{contact}+\mathcal{E}_\mathrm{slip}+\mathcal{E}_\mathrm{temporal}$)  &  \textbf{7.5} & \textbf{266} & \textbf{74.9} \\ \bottomrule 
\end{tabular}
\end{adjustbox}
\end{table}
%
%
%
\begin{figure*}[htbp]
    \mbox{} \hfill
    \centering
    \includegraphics[width=\textwidth]{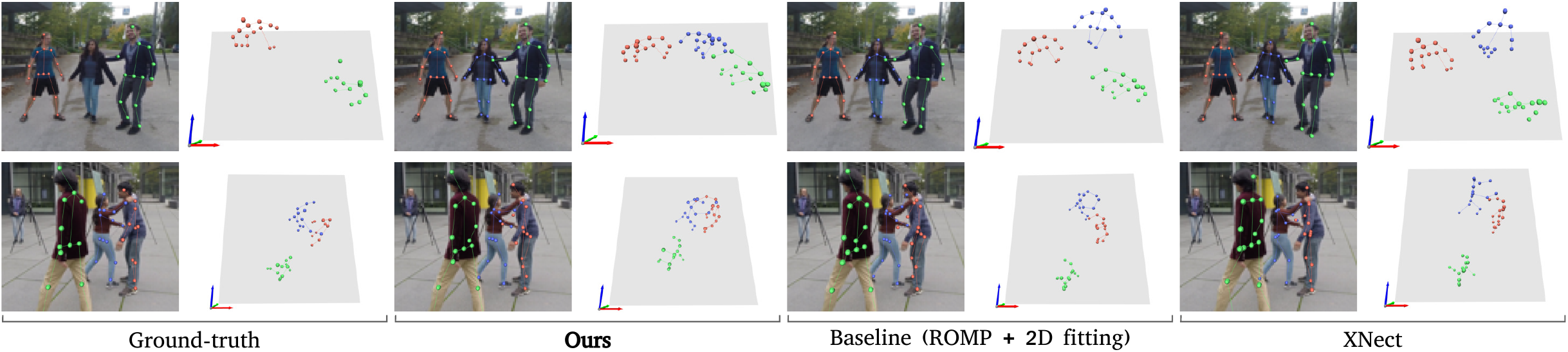}
    \hfill \mbox{}
    \caption{\label{fig:qualitative_results_and_gt}%
        Comparisons of predictions from our method with other approaches. 
        \new{%
        Compared to XNect and our baseline, our method is the only one that is able to estimate the person scale.  Therefore, it predicts human positions in a more coherent way even for people of smaller height.
        3D human poses are shown in the image plane (left) and top view (right). 
        The ground-truth pose is not available for all the subjects in the dataset.
        Digital zoom is recommended.
        }
    }
\end{figure*}
%
\subsection{Ablation Study} \label{sec:ablation}
In this section, we perform additional evaluations of the different components of our method. The results \finalrev{on MuPoTs-3D} are shown in Tables~\ref{tab:ablation_energy} and \ref{tab:ablation_data}. First, we evaluate the influence of the energy terms of \new{the first part of} our optimization framework.
The energy term $\mathcal{E}_\mathrm{depth}$ provides \new{essential} information to disambiguate depth and scale, which contributes to improving the position estimation.
The flexibility provided by the person scale factor can be detrimental to the overall accuracy of the method if no constraints are imposed on it. This can be seen in the second row of Table~\ref{tab:ablation_energy}, without $\mathcal{E}_\mathrm{scale}$.
\new{By constraining our predictions to remain close to the original estimates from ROMP, our method enforces the final estimates to be valid and prevents them from collapsing, as shown in the results without $\mathcal{E}_\mathrm{smpl}$.
}
Finally,
\new{%
$\mathcal{E}_\mathrm{speed}$ is relevant for reducing jitter and the silhouette term provides beneficial contributions to all the metrics. With all the energy terms, our method is stable and precise in estimating 3D position and pose.
}
%
\begin{table}[]
\caption{\label{tab:ablation_energy}%
    Ablation study for different energy terms.
    \new{%
    Without the proposed depth and scale terms, the global position in 3D cannot be precisely recovered, \ie, AP${}_{25}^{root}$ drops from $62.3$ to $47.4\%$ and to $22.2\%$, respectively.
    The SMPL term is critical for enforcing valid estimates, and the speed term contributes to reducing the jitter.
    The silhouette term provides consistent improvements in all the metrics. 
    }
}
\begin{adjustbox}{width=0.4\textwidth,center}
\begin{tabular}{@{}lcccc@{}}
\toprule
Experiment & Jitter~$\downarrow$ & MRPE~$\downarrow$ & AP${}_{25}^{root}$~$\uparrow$ & 3DPCK$_{3d}$~$\uparrow$ \\ \midrule
w/o $\mathcal{E}_\mathrm{depth}$  & 7.8 & 284 & 47.4 & \textbf{75.5} \\ 
w/o $\mathcal{E}_\mathrm{scale}$  & 7.7 & 541 & 22.2 & 68.9 \\ 
w/o $\mathcal{E}_\mathrm{smpl}$  & 8.0 & 674 & 11.5 & 56.3 \\ 
w/o $\mathcal{E}_\mathrm{speed}$  & 8.9 & \underline{269} & \textbf{63.6} & 74.8 \\ 
w/o $\mathcal{E}_\mathrm{silhouette}$  & \underline{7.6} & 270 & 62.0 & 74.7 \\ \midrule  
\textbf{Ours} (full) & \textbf{7.5} & \textbf{266} & \underline{62.3} & \underline{74.9} \\ 
\bottomrule
\end{tabular}
\end{adjustbox}
\end{table}
%
\begin{table}[]
\caption{\label{tab:ablation_data}%
Our results considering different models for 2D pose and monocular depth estimation.
We observe that \new{the human position estimation from} our method benefits directly from advances in the monocular depth estimation
\new{%
when comparing MiDaS v2.1~\cite{ranftl2020towards} and DPT-Large~\cite{ranftl2021vision}.
}
}
\begin{adjustbox}{width=0.45\textwidth,center}
\begin{tabular}{@{}llccc@{}}
\toprule
2D Pose Model & Depth Model & MRPE~$\downarrow$ & AP${}_{25}^{root}$~$\uparrow$ &  3DPCK$_{3d}$~$\uparrow$ \\ \midrule
AlphaPose & MiDaS v2.1 & 278  & 55.8  & \textbf{75.7}  \\ 
AlphaPose & DPT-Hybrid & 276 & 60.8 & 75.0  \\ 
AlphaPose & DPT-Large & \textbf{266} & \textbf{62.3} & 74.9  \\ 
HRNet     & DPT-Large & 304 & 54.9 & 72.7 \\ \bottomrule  
\end{tabular}
\end{adjustbox}
\end{table}
%
%
Since our method relies
\new{%
on off-the-shelf predictors as input, we also provide a concise evaluation considering two different 2D pose and three different depth estimation
}
models from the recent literature. The results in Table~\ref{tab:ablation_data} show that the influence of the depth estimation models is relatively small; however, the best performing model is the most recent transformer architecture, which suggests that our approach directly benefits from improved monocular depth estimations.
\new{%
Regarding 2D pose estimation, HRNet~\cite{wang2020deep_HRNet} performed worse than AlphaPose, since HRNet relies on person detection as a first step, which makes it susceptible to detection failures.
}
%
%
\begin{figure*}[htbp]
    \centering
    \mbox{} \hfill
    \includegraphics[width=0.99\textwidth]{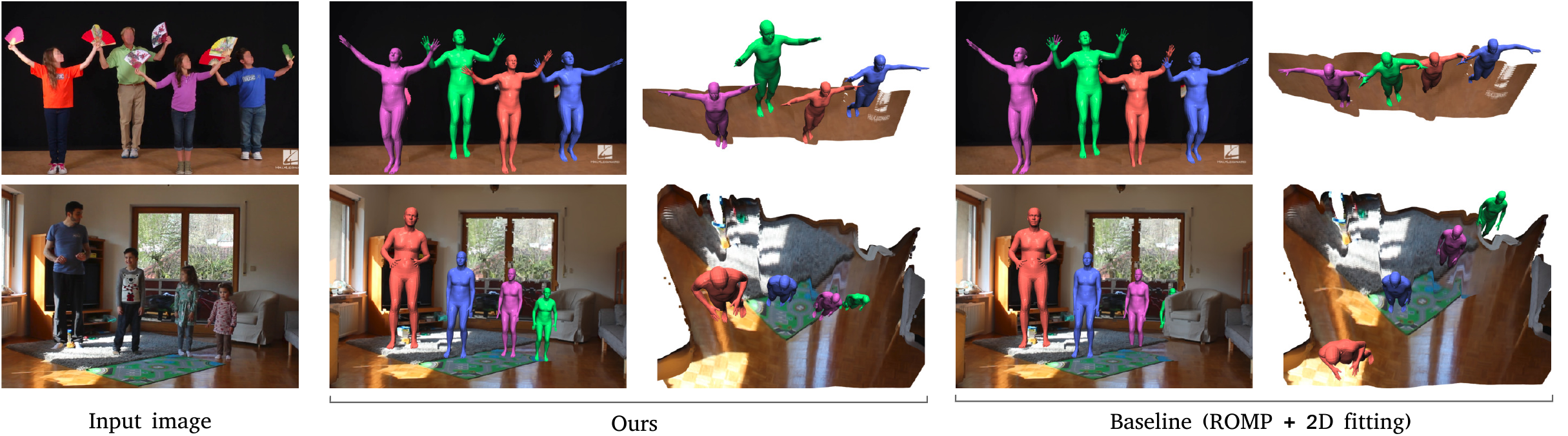}
    \hfill \mbox{}
    \caption{\label{fig:qualitative_results_internet}%
        3D Human poses estimated by our method from Internet videos.
        \new{The baseline method can correctly localise the persons in the image plane, but fails drastically in positioning the characters in 3D}.
        Note \new{from our method} the correct character order along the depth channel \new{and the correctly estimated scale for each person}. 
        Digital zoom is recommended. 
        %
        %
    }
\end{figure*}
%
\begin{figure}[htbp]
    \centering
    \mbox{} \hfill
    \includegraphics[width=0.465\textwidth]{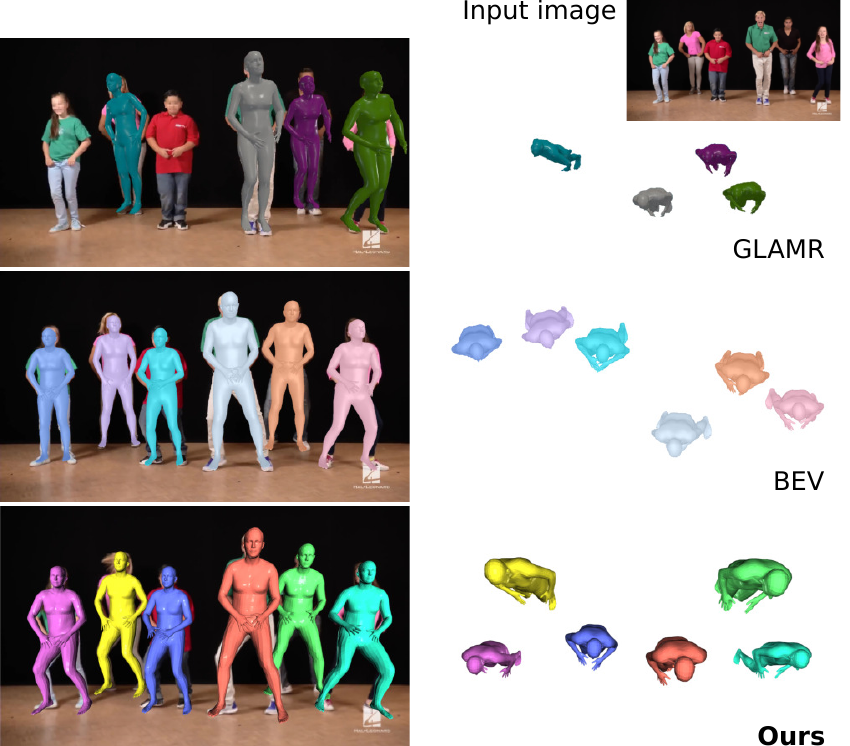}
    \hfill \mbox{}
    \caption{\label{fig:comparison_bev_glamr}%
        \finalrev{Our results compared to BEV~\cite{sun2022putting} and GLAMR~\cite{yuan2022glamr} on a scene with people of different sizes.}
    }
\end{figure}
%
%
\begin{figure}[htbp]
    \centering
    \mbox{} \hfill
    \includegraphics[width=0.475\textwidth]{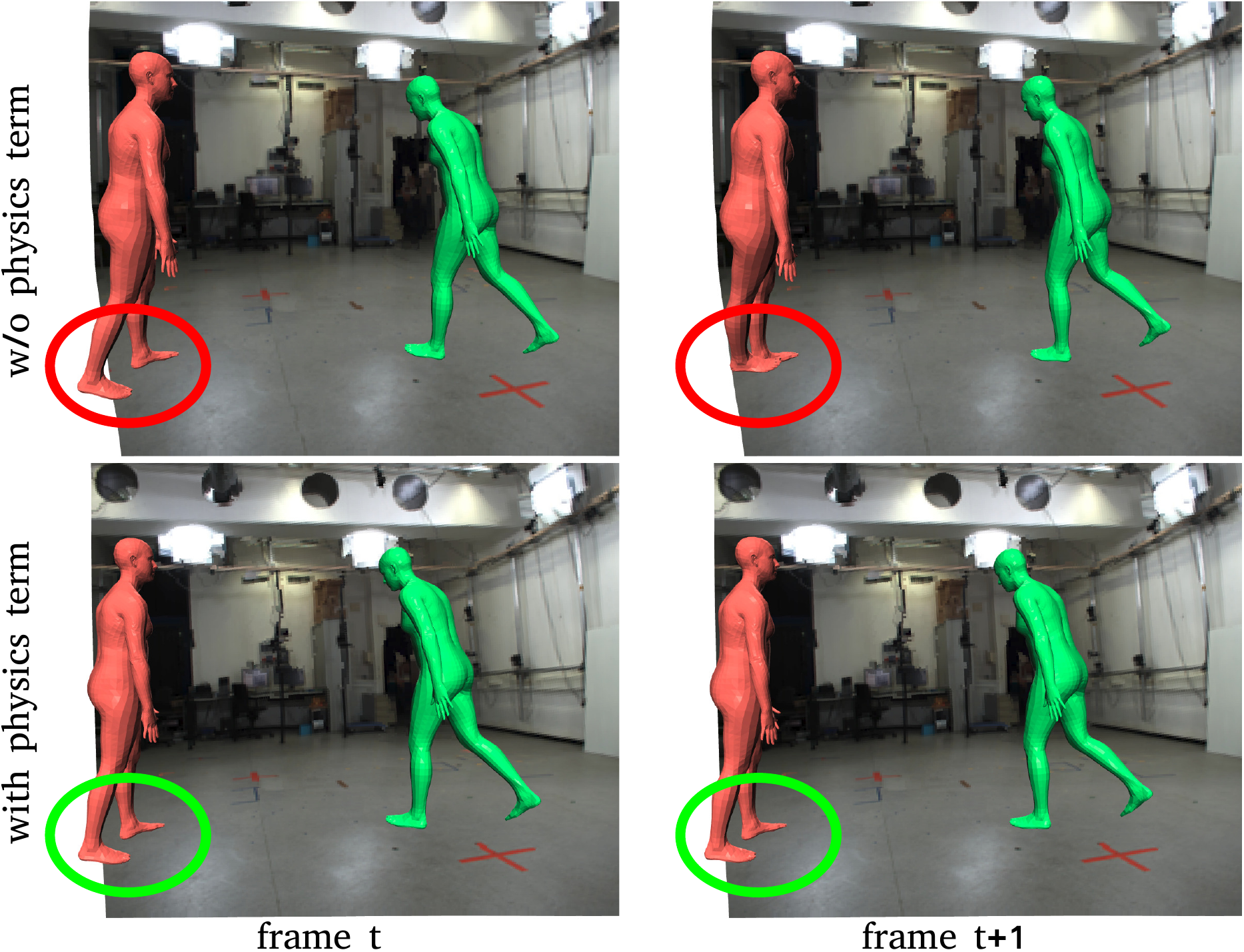}
    \hfill \mbox{}
    \caption{\label{fig:qualitative_results_foot_slidding}%
        The effect of the physical constrains imposed by the estimated geometry in our predictions. The results without $\mathcal{E}_\mathrm{contact}$ and $\mathcal{E}_\mathrm{slip}$ (top) contain more foot sliding artifacts than our results with physical constrains (bottom). 
    }
\end{figure}
%
%
\subsection{Qualitative Results} \label{sec:qualitative}
Figure~\ref{fig:qualitative_results_and_gt} provides  additional qualitative results with predictions from  our method in 3D coordinates, alongside the ground  truth pose. 
We compare our method with XNect~\cite{XNect_SIGGRAPH2020} and ROMP~\cite{ROMP}.
We can see that predictions from ROMP do often not correspond to the correct position of the humans in the scene, \new{since it is not able to estimate the correct person scale}.
For XNect, we can observe that it also fails to recover the correct scale of the person, which can be observed from the top view. 
On the other hand, our approach can predict a 3D pose that corresponds to the ground truth human annotation and is coherently positioned in 3D coordinates. 
\finalrev{We also compare our method with GLAMR~\cite{yuan2022glamr} and BEV~\cite{sun2022putting} in Figure~\ref{fig:comparison_bev_glamr}. GLAMR fails to track all the persons in the scene and BEV fails to predict coherent human positions. More qualitative comparisons are in the supplementary video.}
%
%
%
\par
Our method has the advantage of jointly estimating the humans and the scene point cloud, which can be further used to impose physical constrains in the estimated humans over time. The effect of these constraints can be visually seen in  Figure~\ref{fig:qualitative_results_foot_slidding}, where we show a sequence of a person standing on the floor. 
In the top row, where no physical constraints were applied, we can observe that the right foot 
oscillates drastically from one frame to another. 
When the physical constraints are applied (the bottom row), this artifact is drastically reduced, and the right foot stays still in contact with the ground. 
\par
\begin{figure}[htbp]
    \centering
    \mbox{} \hfill
    \includegraphics[width=0.475\textwidth]{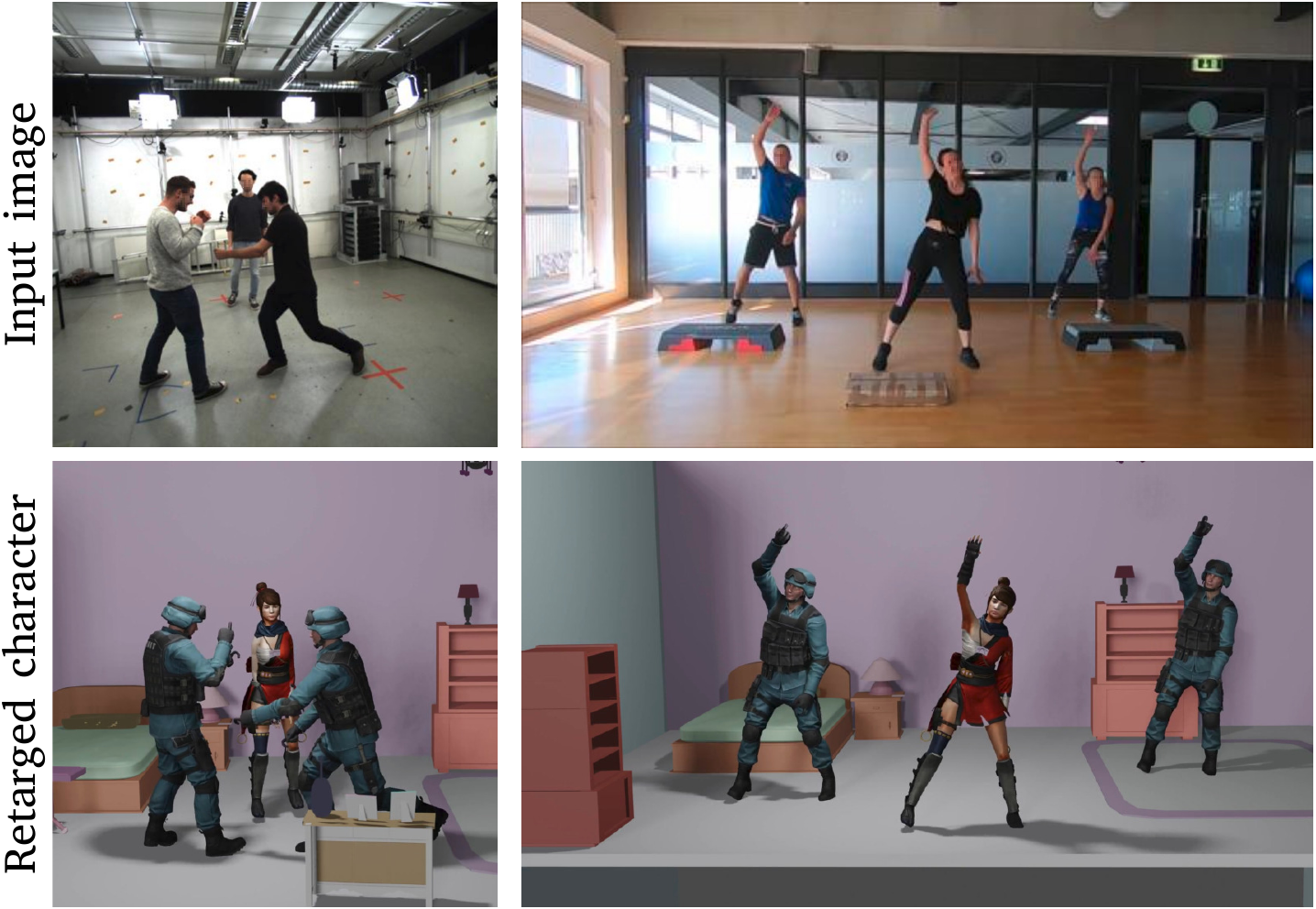}
    \hfill \mbox{}
    \caption{\label{fig:qualitative_characters}%
        Our method can be directly used to drive virtual characters or animate avatars in augmented reality applications (bottom row) from monocular RGB videos. 
        Note the correct character order along the depth channel. 
        Thanks to our physical plausibility constraints, barely any foot-floor penetrations or foot sliding are observed in the animations; see the video. 
        %
    }
\end{figure}

Since our method does not require any specific training procedure and rely on multiple predictions from models trained on a large corpus of data, our approach automatically generalizes well for in-the-wild and Internet videos, as can be seen in Figure~\ref{fig:qualitative_results_internet} 
and can be directly used to drive virtual characters from monocular RGB videos; see Figure~\ref{fig:qualitative_characters}. 
%
%

\section{Discussion}\label{sec:discussion} 
Our method achieves low reconstruction errors, because it can successfully leverage multi-modal inputs to disambiguate the relative depths between humans and human scales better than previous works. 
Moreover, our results evince significantly less jitter and foot-floor penetrations than the evaluated baselines for multi-human 3D pose estimation and the ablative study confirms that all components of the method contribute to the final accuracy. 
We have demonstrated that the recovered 3D human motions can be applied for virtual character animation, as one potential application among the many others. 
\par
\noindent\textbf{Limitations and Possible Extensions.} 
Although our method outperforms competing methods and 
makes a step forward in monocular multi-human 3D motion capture, 
it has several limitations caused by the severe ill-posedness of our monocular setting. All these limitations open possibilities for future extensions and follow-up works as described in the following. \par
%
%
First, our approach relies on multiple inputs from pre-trained models (depth maps and 2D body joints) 
%
and, therefore, could also be negatively affected by the output of those methods; for example if the estimated depth maps contain significant artefacts (\textit{e.g.,} when obtained on our-of-distribution environments). 
On the other hand, this implies that the performance of our approach has the potential to keep increasing in the future with the progress in related fields (\textit{cf.}~Table \ref{tab:ablation_data}). 
%
\par
Our method also requires that people \finalrev{are entirely visible in most of the frames and} move in the scene.
Otherwise, the setting becomes degenerate, and we do not get enough cues for accurate reconstruction. 
Even though we mitigate artefacts that appear as violations of  physical laws by geometric terms, some minor ones of this type remain. 
Further improvements can be attained by methods  explicitly modelling physical laws as in single-human 3D motion capture \cite{PhysCapTOG2020,  PhysAwareTOG2021, Xie2021}. 
%
%
\par
Moreover, while the static camera assumption is practical, 
it is also very challenging, and a moving camera could provide additional 3D  reconstruction cues. 
%
%
Finally, the proposed approach is an optimization method that can efficiently process an entire video sequence and extract relevant information about the scene from all frames globally. 
%
However, due to this characteristic, the method in its current version does not allow real-time applications. 
%


\section{Conclusion}\label{sec:conclusions}
%

%
We present a new holistic approach for multi-human 3D motion capture from a single static monocular RGB camera. 
Our core statement---that the synergy between multi-modal inputs and priors can significantly boost the 3D reconstruction accuracy in this challenging setting---is confirmed by extensive experiments in which we set a new state of the art on commonly used benchmarks. 
Moreover, as expected, we confirm that the constraints from the scene point clouds steadily boost the accuracy of the final 3D poses. 
%
%
%
%
%
%
%
%
%
Qualitatively, our reconstructions evince substantially fewer artefacts (such as jitter and foot-floor penetrations), enabling exciting downstream applications  
such as motion re-targeting for virtual characters. 
%
%
We believe that the proposed holistic approach for multi-human 3D motion capture 
can be extended in many useful ways, and we will be excited to see follow-ups. 
%

\section{Acknowledgments}
This work was funded by the ERC Consolidator Grant 4DRepLy (770784) and the German Science Foundation (DFG) under Grant No. 468670075.

{\small
\bibliographystyle{ieee_fullname}
\bibliography{references}
}

\end{document}